\documentclass[11pt]{cambrian}
\usepackage[utf8]{inputenc}
\usepackage[
    style=numeric-comp,
    doi=true,
    url=false,
    uniquename=false,
    giveninits=true,
    natbib,
    backend=biber]{biblatex}
\usepackage{colortbl}

\usepackage[utf8]{inputenc} 
\usepackage[T1]{fontenc}    
\usepackage{hyperref}
\usepackage{url}            
\usepackage{booktabs}       
\usepackage{amsfonts}       
\usepackage{nicefrac}       
\usepackage{microtype}      
\usepackage{xcolor}
\definecolor{bluelink}{RGB}{0,113,188}
\definecolor{greenlink}{RGB}{0,188,113}
\hypersetup{
    colorlinks=true,%
    citecolor=green!93!black,%
    filecolor=redlink,%
    linkcolor=black,%
    urlcolor=bluelink
}
\usepackage{tabularx}
\usepackage[most]{tcolorbox}
\usepackage{amsmath}
\usepackage{multirow}
\usepackage{array}
\usepackage{caption}
\usepackage{wrapfig}
\usepackage{enumitem}
\usepackage{tikz}
\usepackage{lipsum}
\usepackage{subcaption}
\usepackage{multirow} 
\usepackage{adjustbox}

\usepackage{amssymb}
\usepackage{unicode}  
\usepackage[capitalize]{cleveref} 

\usepackage{pgf}
\usepackage{colortbl}

\usepackage{tipa}

\usepackage{tcolorbox}
\usepackage{rotating}
\usepackage[abs]{overpic}
\usepackage{makecell}
\usepackage{longtable}

\usepackage{tocloft}  

\usepackage[english]{babel}
\usepackage{csquotes}
\usepackage{listings}
\definecolor{codekeyword}{rgb}{0.0, 0.0, 0.5}   
\definecolor{codecomment}{rgb}{0.0, 0.5, 0.0}   
\definecolor{codestring}{rgb}{0.56, 0.0, 1.0}   

\usepackage{cuted}
\usepackage[ruled,vlined]{algorithm2e}
\definecolor{lightblue}{RGB}{220,235,250}
\definecolor{lightgreen}{RGB}{220,235,250}

\lstdefinestyle{pythonstyle}{
    language=Python,                          
    basicstyle=\ttfamily\small,               
    keywordstyle=\color{codekeyword}\bfseries,
    commentstyle=\color{codecomment}\itshape, 
    stringstyle=\color{codestring},           
    showstringspaces=false,                   
    breaklines=true,                          
    tabsize=4,                                
    numbers=none,                             
    frame=none,                               
    backgroundcolor=\color{white},            
    captionpos=b,                             
    morekeywords={self, __init__, __name__, __main__}, 
}

\newcommand{\eg}{{e.g.}}
\newcommand{\ie}{{i.e.}}

\title{\center EVOLVE-VLA: Test-Time Training from Environment Feedback for Vision-Language-Action Models}

\renewcommand{\thefootnote}{\fnsymbol{footnote}}

\author{Zechen Bai,
    Chen Gao,
    Mike Zheng Shou\footnote{Corresponding Author}\\
    Show Lab, National University of Singapore \\
    \url{https://showlab.github.io/EVOLVE-VLA}
}

\begin{abstract}
Achieving truly adaptive embodied intelligence requires agents that learn not just by imitating static demonstrations, but by continuously improving through environmental interaction, which is akin to how humans master skills through practice.
Vision-Language-Action (VLA) models have advanced robotic manipulation by leveraging large language models, yet remain fundamentally limited by Supervised Finetuning (SFT): requiring hundreds of demonstrations per task, rigidly memorizing trajectories, and failing to adapt when deployment conditions deviate from training.
We introduce EVOLVE-VLA, a test-time training framework enabling VLAs to continuously adapt through environment interaction with minimal or zero task-specific demonstrations.
The key technical challenge is replacing oracle reward signals (unavailable at test time) with autonomous feedback.
We address this through a learned progress estimator providing dense feedback, and critically, we design our framework to ``tame'' this inherently noisy signal via two mechanisms:
(1) an accumulative progress estimation mechanism smoothing noisy point-wise estimates, and (2) a progressive horizon extension strategy enabling gradual policy evolution.
EVOLVE-VLA achieves substantial gains: +8.6\% on long-horizon tasks, +22.0\% in 1-shot learning, and enables cross-task generalization—achieving 20.8\% success on unseen tasks without task-specific demonstrations training (vs. 0\% for pure SFT).
Qualitative analysis reveals emergent capabilities absent in demonstrations, including error recovery and novel strategies.
This work represents a critical step toward VLAs that truly learn and adapt, moving beyond static imitation toward continuous self-improvements.
\end{abstract}

\begin{document}
\maketitle
\setcounter{footnote}{0}  
\renewcommand{\thefootnote}{\arabic{footnote}}  

\begin{figure}[t!]
    \begin{center}
        \includegraphics[width=0.97\linewidth]{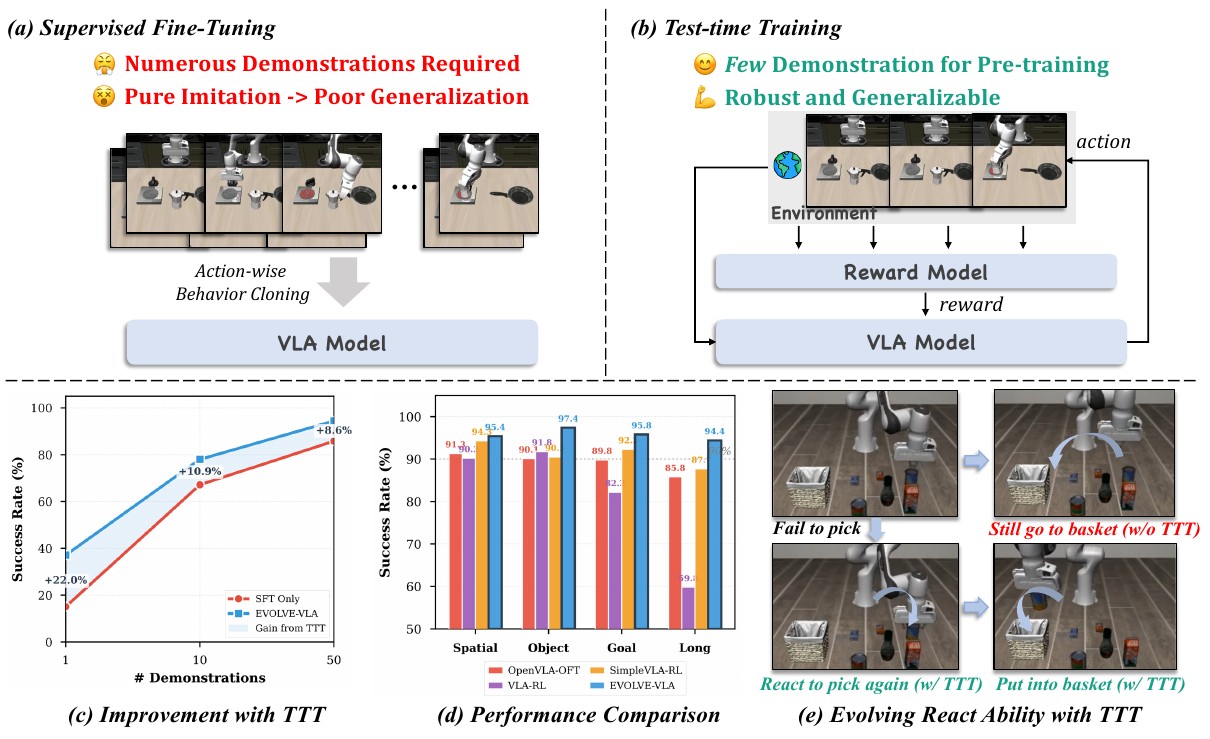}
        \captionof{figure}{
    \textbf{(a)}
    Traditional supervised finetuning paradigm for training VLA model requires numerous demonstration data and risks rigidly cloning trajectories in the training data.
    \textbf{(b)}
    Our test-time training framework requires only few demonstrations (even none) for pre-training and can evolve in the deployment environment.
    In \textbf{(c)} and \textbf{(d)}, we show that our method achieves performance gains across various number of demonstrations and different task suites.
    Notably, for the first time, we observe zero-shot cross-task generation with the test-time-training framework.
    \textbf{(e)} illustrates the ability of recovering from mistakes evolved during TTT.
            }
        \label{fig:teaser}
    \end{center}
    \vspace{-0.3cm}
\end{figure}

\section{Introduction}

How do humans develop manipulation skills? We do not simply watch an expert perform a task once and then flawlessly replicate it. Instead, we learn through practice: attempting the task repeatedly, making mistakes, receiving feedback from the environment, and gradually refining our movements through continued experience. This process of learning by doing, rather than merely learning by watching, is fundamental to how intelligent agents acquire robust and adaptable capabilities in the real world.

Embodied intelligence, the integration of AI technology with robots, has seen remarkable progress in recent years.
Propelled by the capabilities of Large Language Models (LLMs), control policies are rapidly evolving beyond traditional methods toward general Vision-Language-Action (VLA) models~\cite{kim2024openvla,black2024pi0,kim2025fine}, which process multimodal inputs to produce a sequence of actions for completing a given task.
By leveraging the rich semantic priors from LLMs, VLAs demonstrate impressive contextual understanding compared to their predecessors.
However, despite these advances, current VLA training remains fundamentally misaligned with the human learning principle described above: they are trained exclusively through Supervised Fine-Tuning (SFT) on fixed demonstration datasets, learning to imitate expert behavior but lacking mechanism to improve through environmental interaction.

This paradigm of static imitation learning entails two fundamental limitations.
\emph{\textbf{(1) High labor cost.}} As shown in Fig.~\ref{fig:teaser}(a), adapting VLA models to new tasks requires collecting hundreds of demonstrations for supervised fine-tuning (SFT). This cost multiplies linearly with tasks, making it infeasible to scale VLAs to truly general-purpose robots.
\emph{\textbf{(2) Brittle memorization.}}
VLAs optimized through behavior cloning merely imitate demonstrations and struggle to generalize beyond training distribution.
They lack the ability to recover from execution deviations, where a single misstep often leads to complete task failure.
These limitations represent a fundamental misalignment with how adaptive intelligence should operate. We believe that enabling continuous learning from deployment experience is essential for achieving truly general-purpose vision-language-action models.

In this work, we propose EVOLVE-VLA (\textbf{E}fficient \textbf{VLA} \textbf{O}nline \textbf{L}earning \textbf{V}ia \textbf{E}xperience), a test-time training framework that fundamentally shifts how VLAs learn and adapt.
As illustrated in Fig.~\ref{fig:teaser}(b), instead of requiring hundreds of expert demonstrations, our method needs only minimal supervision, a few demonstrations or even none, for lightweight initialization via SFT. The key innovation lies in what happens after this initial pre-training: rather than freezing the policy, we deploy it directly in the target environment where it continues to learn autonomously through active interaction.
The VLA explores the environment, receives feedback, and refines its behavior via online reinforcement learning, mirroring the trial-and-error process through which humans develop manipulation skills.

This paradigm shift addresses both limitations: (1) it dramatically reduces labor costs by replacing extensive demonstrations with autonomous learning, and (2) it enables genuine adaptation rather than memorization, producing policies that recover from errors and discover novel strategies. For example, Fig.~\ref{fig:teaser}(e) shows our model developing error correction capabilities absent from training demonstrations. Beyond improving seen tasks, this approach enables cross-task generalization through self-directed exploration. 

While prior works like SimpleVLA-RL~\cite{li2025simplevla} have explored RL for VLA models, they rely on oracle reward functions (\eg, binary success signals) unavailable at test time.
The central challenge of practical TTT is replacing the oracle with autonomous feedback.
We introduce a learned progress estimator as reward, with the policy optimized via GRPO~\cite{shao2024deepseekmath}.
Unlike sparse success signals, progress-based rewards provide dense, continuous feedback crucial for sample-efficient learning.
However, practical progress estimators are inherently noisy~\cite{zhang2025rewind,gvl,sontakke2023roboclip,ma2023liv}, and errors accumulated over long horizons can mislead the policy.

Our core technical challenge is therefore not to build a perfect estimator, but to successfully ``tame'' this noisy reward signal to make learning possible.
To achieve this, we introduce two key technical contributions.
First, we design an \textit{accumulative progress estimation mechanism} with interval-based sampling, which aggregates and smooths noisy point-wise estimates into a stable, reliable signal.
Second, we propose a \textit{progressive horizon extension strategy} that optimizes the policy with progressively increasing exploration horizon, making the model more resilient to estimation errors by allowing it to first master simpler sub-tasks.
This combined approach not only mitigates the impact of estimation noise but also allows the VLA to effectively utilize the dense, albeit imperfect, reward.

Our framework enables VLA models to perform test-time training using self-generated environmental feedback without oracle rewards.
We validate EVOLVE-VLA on the LIBERO benchmark, achieving substantial gains: +8.6\% on long-horizon tasks, +22.0\% in 1-shot learning, and cross-task transfer (0\% → 20.8\% on unseen tasks through autonomous adaptation).
Qualitative analysis reveals emergent capabilities absent from demonstrations, including error recovery and novel strategies.
These results validate that test-time training represents a paradigm shift toward adaptive embodied agents—a critical step toward truly general-purpose VLA systems.
Our contributions include:
\begin{itemize}
    \item We propose EVOLVE-VLA, a \textbf{test-time training framework} that enables VLAs to continuously adapt through autonomous interaction, addressing the brittleness and scalability limitations of static SFT. 
    
    \item We tackle the central challenge of absence of oracle rewards by introducing a learned progress estimator. Critically, we develop techniques to \emph{``tame'' inherently noisy reward signals}, making practical test-time training feasible.
    
    \item We introduce two key innovations: (1) an \textbf{accumulative progress estimation mechanism} that smooths noisy estimates into stable signals, and (2) a \textbf{progressive horizon extension strategy} enabling gradual policy evolution, proving effective for long-horizon tasks.
    
    \item We demonstrate strong results: \textbf{+8.6\%} on long-horizon tasks, \textbf{+22.0\%} in 1-shot learning, and pioneering \textbf{zero-shot cross-task generalization} (0\% → 20.8\%) through test-time adaptation alone. Our analysis reveals emergent skills like error recovery arising from autonomous exploration.
\end{itemize}

\section{Related Work}

\noindent\textbf{Vision-Language-Action Models.}
Recent advances in Vision-Language-Action (VLA) models~\cite{zhao2025cot,black2024pi0,ding2024quar,kim2024openvla,qu2025spatialvla,wen2025tinyvla,kim2025fine,wen2025diffusionvla} aim to equip embodied agents with the ability to perceive, reason, and act upon multimodal inputs. 
Early works like RT~\cite{brohan2022rt} and Octo~\cite{team2024octo} investigate how to connect the power of large models with the interactive nature of embodied environments, paving the way toward generalist robot manipulation. OpenVLA~\cite{kim2024openvla} presents an open-source VLA model fine-tuned across multiple manipulation tasks, aiming to standardize evaluation and promote reproducible research. OpenVLA-OFT~\cite{kim2025fine} further proposes parallel decoding, action chunking, and a continuous action representation to improve performance. $\pi_0$~\cite{black2024pi0} introduces a VLA flow model by a continuous flow-based architecture. The approach demonstrates strong generalization across diverse robot manipulation tasks and sets a new direction for flow-based embodied reasoning.

Some works focus on improving the efficiency of VLA model. TinyVLA~\cite{wen2025tinyvla} designs a lightweight VLA for robotic manipulation, which employs parameter sharing and distillation to retain performance under limited data. 
Recent works~\cite{bi2025vla,yang2025bitla,guo2025omnivla} also investigate how to involve tactile modality in VLA models.
However, previous methods rely heavily on imitation learning with numerous manual-collected data, leading to labor-cost and poor generalization models, especially when meeting the new tasks and environments. 

\noindent\textbf{RL Fine-Tuning for VLA Models.}
With the recent advances of RL post-training in LLMs~\cite{touvron2023llama,brown2020language} and MLLMs~\cite{llava,wang2024qwen2vl}, some studies have begun to explore RL post-training for VLA models.
For example, iRe-VLA~\cite{guo2025improving} explores how online RL can enhance pretrained VLA models by allowing continual improvement through interaction. VLA-RL~\cite{lu2025vla} introduces a trajectory-level RL formulation for VLA training. OctoNav~\cite{gao2025octonav} investigates how GRPO-like RL training can improve VLA reasoning ability in embodied navigation. SimpleVLA-RL~\cite{li2025simplevla} and $\pi_{RL}$~\cite{chen2025pirl} explore RL fine-tuning for autoregressive and flow-based VLAs, respectively. RL4VLA~\cite{liu2025can} systematically studies different RL policies and the impact of RL fine-tuning across diverse visual, semantic, and execution dimensions.
Although these works have explored RL post-training strategies for VLA models, they still assume access to Ground-Truth (GT) information during the RL training phase, such as whether a trajectory succeeds or fails. 
However, at test time, such GT supervision signals are unavailable.
To address this, we propose a test-time training framework that enables the model to adapt without relying on GT feedback.

\noindent\textbf{Concurrent Work: $\pi^*0.6$.}
Concurrent to our work, Physical Intelligence recently released $\pi^*0.6$~\cite{pistar06}, a vision-language-action model that learns from autonomous experience using their Recap method (RL with Experience \& Corrections via Advantage-conditioned Policies). Our work shares a similar motivation and spirit with $\pi^*0.6$ in addressing a fundamental limitation of VLA models trained purely on demonstration data: the inability to handle compounding errors and improve from deployment experience.

The concurrent emergence of both works from academia and industry highlights a growing recognition that \textit{experience-based reinforcement learning is essential for VLA models to move beyond behavior cloning}. Both approaches demonstrate that achieving reliable and robust performance requires learning from the robot's own experience rather than solely imitating expert demonstrations. We submitted EVOLVE-VLAbefore the release of $\pi^*0.6$, representing pioneering academic work in this direction. To foster further research and democratize access to this paradigm, we commit to releasing our full training and inference codebase upon publication.

\section{Method}

\begin{figure*}[ht]
  \centering
  \includegraphics[width=1.0\linewidth]{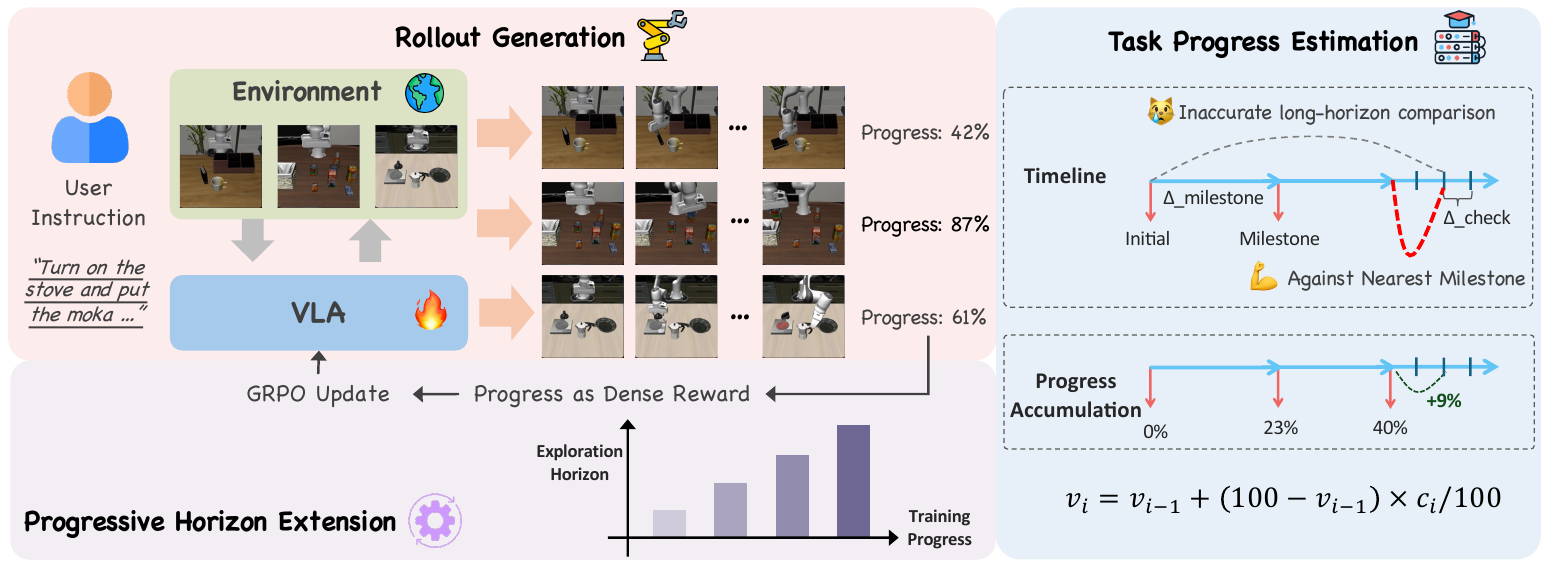}
  \caption{
  \textbf{Framework overview.}
  In the test-time training phase, a VLA model interacts with the environment and generates diverse rollout trajectories.
  A task progress estimation module assign a progress for each rollout, which will be utilized as the reward for GRPO optimization.
  Task progress estimation employs an accumulative strategy that produces clean, stable and smooth reward.
  The training strategy undergoes a progressive horizon extension schedule, enabling a learning curriculum.
  }
  \label{fig:method}
\end{figure*}

\subsection{Task Definition}
\label{sec:task_definition}

We formulate the robotic manipulation task as a Markov Decision Process (MDP) $\mathcal{M} = (\mathcal{S}, \mathcal{A}, P, \mathcal{R}, \gamma)$, where $\mathcal{S}$ is the state space, $\mathcal{A}$ is the action space, $P$ represents transition dynamics, $\mathcal{R}$ is the reward function, and $\gamma \in [0,1)$ is the discount factor.
At timestep $t$, the state $s_t = (o_t^{\text{vis}}, o_t^{\text{prop}}, l_{\text{task}})$ consists of visual observation $o_t^{\text{vis}}$, proprioceptive state $o_t^{\text{prop}}$, and task instruction $l_{\text{task}}$.

A VLA policy $\pi_\theta: \mathcal{S} \rightarrow \Delta(\mathcal{A})$ maps states to action distributions. Following modern VLA architectures~\cite{black2024pi0,kim2024openvla}, we adopt action tokenization where continuous robot actions $a_t \in \mathbb{R}^d$ are discretized into tokens. The policy autoregressively generates action token sequences $\mathbf{a} = (a_1, \ldots, a_T)$ with probability $\pi_\theta(\mathbf{a} \mid s_t) = \prod_{k=1}^{T} \pi_\theta(a_k \mid s_t, a_{<k})$.
A trajectory $\tau = \{(s_0, a_0), \ldots, (s_H, a_H)\}$ is generated through closed-loop interaction: the policy outputs actions, the environment transitions based on physical dynamics, and updated observations feed back into the policy until task completion or maximum horizon $H$.

\subsection{Test-time Training Framework}
\label{sec:ttt_framework}

During deployment, a VLA model pretrained via SFT on expert demonstrations encounters novel scenarios that differ from its training distribution. Traditional SFT models, which learn purely through imitation, lack the mechanism to adapt to these out-of-distribution states. Our goal is to enable the VLA to continue learning at test-time by leveraging online interaction with the environment.

Test-time training (TTT) requires two key components: (1) the ability to actively interact with the environment to generate diverse rollouts, and (2) a feedback signal to evaluate and improve these rollouts. We achieve this through online reinforcement learning, where the policy is iteratively refined based on rewards obtained from environment interaction.
Fig.~\ref{fig:method} shows the overview of our TTT framework.

\subsubsection{Online Reinforcement Learning}
\label{sec:online_rl}

\textbf{Interactive Rollout Generation.}
For a given task, we generate multiple diverse trajectories by sampling from the policy's action token distribution with temperature $T > 1$. Specifically, starting from initial state $s_0$, at each timestep $t$, the policy outputs action token probabilities and samples an action $a_t$ from the distribution. This action is executed in the environment, producing a new state $s_{t+1}$. This closed-loop interaction continues until the estimated task progress exceeds a threshold (indicating completion) or the maximum horizon $H_{\text{max}}$ is reached, yielding a trajectory $\tau_i = \{(s_0, a_0), \ldots, (s_H, a_H)\}$. By sampling $G$ trajectories $\{\tau_i\}_{i=1}^G$ with different random seeds, we explore diverse solution strategies.

\textbf{Environment Feedback.}
Each trajectory receives a reward $R_i$ that evaluates its quality. This reward signal, which we detail in \S\ref{sec:progress_reward}, serves as the supervisory feedback guiding policy improvement. Unlike SFT which only learns from successful demonstrations, the reward signal provides differential feedback, distinguishing better trajectories from worse ones and enabling the model to discover and reinforce effective behaviors through trial and error.

\textbf{Policy Update.}
We employ Group Relative Policy Optimization (GRPO)~\cite{shao2024deepseekmath} to update the policy. GRPO normalizes trajectory rewards within each batch to compute advantages and applies PPO-style clipping for stable updates, without requiring a separate value network.

\subsubsection{Task Progress Estimation}
\label{sec:progress_reward}

A critical challenge for test-time training is the absence of oracle reward signals (e.g., ground-truth success indicators from simulators) that are available during training in simulator but unavailable at deployment. We address this by learning a reward function based on \textit{task progress}: an estimate of how much of the task has been completed.

\textbf{Task Progress as Reward Function.}
Progress-based rewards offer several advantages over binary success signals. First, they are \textit{dense}: progress can be estimated at any point during execution, providing continuous feedback even for failed attempts. This density is crucial for sample-efficient learning, especially in long-horizon tasks where successful rollouts may be rare initially. Second, progress is a more general, grounded concept than task-specific metrics or black-box reward scores, making it applicable across diverse manipulation tasks.

\textbf{Task Progress as Termination Condition.}
Beyond providing rewards, task progress estimation also determines when to terminate rollouts.
When estimated progress exceeds a predefined threshold, the rollout stops as the task is deemed complete; otherwise, execution continues until maximum horizon $H_{\text{max}}$.
This dual-purpose usage imposes stringent requirements on the estimator: it must be (1) \textit{computationally efficient}, as it is queried frequently (every $\Delta_{\text{check}}$ steps) to detect completion in real-time, and (2) \textit{temporally smooth and consistent}, as erratic estimates can cause premature termination (stopping promising trajectories early) or delayed termination (wasting computation on completed tasks).
While noisy rewards can be mitigated through averaging during policy learning, a single erroneous termination decision can truncate an entire trajectory.
Therefore, stabilizing the progress signal is essential not just for learning efficiency, but for correct rollout execution.

\textbf{Vanilla Progress Estimation.}
We employ a foundation critic model, VLAC~\cite{zhai2025vlac}, which takes two images and task instruction as input and output a critic value.
A positive value indicates how much the second image progresses the task compared to the first image.
A negative value vice versa.
Specifically, given a trajectory $\tau = \{(s_0, a_0), \ldots, (s_H, a_H)\}$, we compute the reward as $R_i = \text{Critic}(o_0, o_H, l_{\text{task}})$, where $o_0$ and $o_H$ are the initial and final observations of the trajectory, and $l_{\text{task}}$ is the task instruction.
The estimated reward is then normalized to $[0, 1]$ to serve as the trajectory reward $R_i$ for GRPO.

\subsection{Accumulative Progress Estimation}
\label{sec:accumulative}

While the progress critic provides dense feedback, we observe that it can be noisy and inconsistent, especially for long-horizon tasks involving multiple sub-goals. A single frame-pair comparison may be misled by superficial visual changes or fail to capture intermediate progress.
As discussed in \S\ref{sec:progress_reward}, this noisy estimation can negatively affect both reward feedback and rollout termination.

To address these challenges, we introduce an \textit{accumulative progress estimation} mechanism.
Our key insight is inspired by a slow-fast philosophy: instead of comparing the final state to the very beginning (which becomes unreliable for long trajectories), we maintain \textit{milestone frames} at regular intervals and compute progress incrementally.

\textbf{Interval-Based Milestone Sampling.}
We define a sampling interval $\Delta_{\text{milestone}}$ (e.g., 64 timesteps). During rollout, we maintain a list of milestone frames $\mathcal{F}_{\text{milestone}} = \{f_0, f_{\Delta_{\text{milestone}}}, f_{2\Delta_{\text{milestone}}}, \ldots\}$ that captures the trajectory's evolution at a coarse granularity. These milestones serve as reference points for measuring progress.

\textbf{Incremental Progress Computation.}
At a finer granularity (every $\Delta_{\text{check}}$ steps, where $\Delta_{\text{check}} < \Delta_{\text{milestone}}$), we query the critic to estimate progress relative to the most recent milestone. Specifically, at timestep $t$, we compute:
\begin{equation}
c_t = \text{Critic}(f_{\lfloor t / \Delta_{\text{milestone}} \rfloor \cdot \Delta_{\text{milestone}}}, o_t),
\end{equation}
where $c_t \in [-100, 100]$ represents the incremental progress from the last milestone to the current state. When $t$ reaches a new milestone ($t \bmod \Delta_{\text{milestone}} = 0$), we append $o_t$ to $\mathcal{F}_{\text{milestone}}$ and store $c_t$ in the critic history.

\textbf{Accumulative Value Aggregation.}
Given a sequence of incremental critic values\\
$\{c_{\Delta_{\text{milestone}}}, c_{2\Delta_{\text{milestone}}}, \ldots, c_{k\Delta_{\text{milestone}}}\}$ collected at milestones, we accumulate them into a progress value $v_t \in [0, 100]$ that estimates task completion percentage:
\begin{equation}
v_{i} = v_{i-1} + (100 - v_{i-1}) \cdot c_{i} / 100, \quad v_0 = 0,
\end{equation}
where $i$ indexes the milestones. This recursive formulation applies a \textit{diminishing returns} principle: positive progress advances the value toward 100 by a fraction of the remaining distance, while negative critics decrease the value proportionally. Critically, adjustments scale with $(100 - v_{i-1})$ (the remaining gap to completion) prevents both overshooting from overly optimistic critics and catastrophic collapse from pessimistic ones.

\begin{algorithm}[t]
\small
\caption{Accumulative Progress Estimation}
\label{alg:accumulative_progress}
\KwIn{Critic model, $\Delta_{\text{milestone}}$, $\Delta_{\text{check}}$, $\tau_{\text{threshold}}$}
\KwOut{Trajectory reward $R_{\text{accum}}(\tau)$}

Initialize milestone frames $\mathcal{F}_{\text{milestone}} \gets [o_0]$\;
Initialize critic history $\mathcal{C} \gets []$\;
Initialize progress values $\mathcal{V} \gets [0]$\;
Initialize $v_{\text{current}} \gets 0$\;

\For{$t = 1$ \KwTo $H_{\text{max}}$}{
    Execute action $a_t$, observe $o_t$\;
    \If{$t \bmod \Delta_{\text{check}} = 0$}{
        $k \gets \lfloor t / \Delta_{\text{milestone}} \rfloor$ \tcp*{Nearest milestone}
        $f_{\text{ref}} \gets \mathcal{F}_{\text{milestone}}[k]$ \tcp*{milestone frame}
        $c_t \gets \text{Critic}(f_{\text{ref}}, o_t)$ \tcp*{Compute incremental progress}
        
        \If{$t \bmod \Delta_{\text{milestone}} = 0$}{
            Append $o_t$ to $\mathcal{F}_{\text{milestone}}$ \tcp*{New milestone}
            Append $c_t$ to $\mathcal{C}$ \tcp*{Store critic value}
            
            \tcp{Accumulate progress value with diminishing returns}
            $v_{\text{current}} \gets v_{\text{current}} + (100 - v_{\text{current}}) \cdot c_t / 100$\;
            Append $v_{\text{current}}$ to $\mathcal{V}$\;
        }
        
        \tcp{Termination check}
        \If{$v_{\text{current}} / 100 > \tau_{\text{threshold}}$}{
            \textbf{break} \tcp*{Task deemed complete}
        }
    }
}

\tcp{Use accumulated progress as reward}
$R_{\text{accum}}(\tau) \gets v_{\text{current}} / 100$\;

\Return{$R_{\text{accum}}(\tau)$}
\end{algorithm}

The full mechanism is shown in Algorithm~\ref{alg:accumulative_progress}.
It effectively smooths the noisy critic: by comparing to recent milestones rather than the distant initial state, we reduce the impact of long-term drift; by applying proportional adjustments rather than raw critic values, we create a more stable learning signal; and by accumulating progress incrementally with diminishing returns, we smooth out local fluctuations.
Such a smoothed reward provides more reliable feedback for the reinforcement optimization.

In addition, this mechanism is also computationally efficient.
Recall that since the progress need to called frequently for  determining rollout termination, at timestep $T$, a naive multi-frame approach would require $T-1$ critic calls to evaluate all pairwise comparisons, whereas our method requires only a single call—comparing the current frame to the nearest milestone.

\subsection{Progressive Horizon Extension}
\label{sec:curriculum}

Long-horizon tasks present a fundamental challenge for test-time training: early in training, the policy is far from proficient and successful task completion is rare, making credit assignment difficult with noisy reward signals.
Simply allowing free exploration until the maximum horizon $H_{\text{max}}$ leads to low-quality trajectories that provide weak learning signals.
Even with our accumulative progress estimation, optimizing over very long horizons from the start can lead to unstable learning dynamics.

To address this, we adopt a \textit{progressive horizon extension} strategy. We divide the training process into stages, where each stage operates with a maximum rollout horizon $H_{\text{max}}$. As training progresses through stages, we gradually increase $H_{\text{max}}$, allowing the policy to first master shorter sub-goals before tackling the complete task. In early stages, the agent focuses on immediate objectives and fundamental manipulation behaviors where the reward signal is cleaner and more direct. As the horizon extends in later stages, the policy learns to chain these behaviors together and reason over longer temporal dependencies, ultimately optimizing complete task execution.

This schedule provides several benefits. First, shorter horizons naturally reduce the accumulation of noise in progress estimation, as fewer milestone comparisons are needed. Second, early success on simpler sub-goals provides positive learning signals that would be absent when optimizing full-length trajectories from scratch. Third, the staged progression allows the policy to build compositional skills, where early stages establish robust primitives, while later stages learn to orchestrate them.

Importantly, progressive learning and accumulative progress estimation are complementary mechanisms. The progressive curriculum addresses \textit{temporal credit assignment}, \ie, determining when and what to learn, by controlling the optimization scope. Accumulative estimation addresses \textit{noisy rewards}, \ie, stabilizing the feedback signal, by aggregating incremental progress. 
Together, they enable robust test-time training on long-horizon manipulation tasks where both challenges are present.

\section{Experiments}
\label{sec:experiments}

\textbf{Benchmark.}
We evaluate our method on the LIBERO benchmark~\cite{liu2023libero}, a widely used simulation benchmark for lifelong learning in robotic manipulation. LIBERO focuses on language-guided manipulation tasks across diverse object types, task specifications, and environments. It consists of four task suites: LIBERO-Spatial, LIBERO-Object, LIBERO-Goal, and LIBERO-Long. Each suite contains 10 tasks, with 50 expert demonstrations per task. We report the average Success Rate (SR) across 50 trials for each task, following the evaluation protocol in previous work~\cite{kim2024openvla,li2025simplevla}.

\textbf{Base Model.}
We apply our method to OpenVLA-OFT~\cite{kim2025fine}, a state-of-the-art autoregressive VLA model that achieves high performance and inference efficiency.
Following prior work~\cite{li2025simplevla}, we adopt the action chunking and parallel decoding designs, while disabling the continuous action regression head, i.e., use discrete action tokens instead.
This would enable action generation compatible with the optimization of reinforcement learning.

\textbf{Reward Model.}
For test-time training, we employ a foundation critic model VLAC~\cite{zhai2025vlac} as the progress estimator. VLAC takes two images and a language instruction as input and outputs a critic value indicating how much the second image represents progress toward task completion compared to the first image.
This foundation model has been pre-trained on large-scale robotic manipulation datasets, demonstrating its ability to estimate task progress across diverse tasks and environments.

\subsection{Main Results}

\begin{table}[!t]
  \centering
  \caption{Main results of different VLA models on LIBERO.}
  \vspace{-6pt}
  \resizebox{.6\linewidth}{!}{
    \begin{tabular}{lccccc}
      \toprule
      \multirow{2}{*}{\textbf{Model}} &
        \multicolumn{5}{c}{\textbf{LIBERO}} \\
      \cmidrule(lr){2-6}
      & \textbf{Spatial} & \textbf{Object} & \textbf{Goal} & \textbf{Long} & \textbf{Avg} \\
      \midrule
      Octo                & 78.9 & 85.7 & 84.6 & 51.1 & 75.1 \\
      OpenVLA             & 84.7 & 88.4 & 79.2 & 53.7 & 76.5 \\
      Nora                & 92.2 & 95.4 & 89.4 & 74.6 & 87.9 \\
      $\pi_0$ + FAST      & 96.4 & 96.8 & 88.6 & 60.2 & 85.5 \\
      $\pi_0$             & 96.8 & 98.8 & 95.8 & 85.2 & 94.2 \\
      UniVLA              & 96.5 & 96.8 & 95.6 & 92.0 & 95.2 \\
      VLA-RL              & 90.2 & 91.8 & 82.2 & 59.8 & 81.0 \\
      SimpleVLA$\dagger$  & 94.3 & 90.5 & 92.3 & 87.7 & 91.2 \\
      \midrule
      OpenVLA-OFT         & 91.3 & 90.1 & 89.8 & 85.8 & 89.2 \\
      \textbf{EVOLVE-VLA}       & 95.4 & 97.4 & 95.8 & 94.4 & 95.8 \\
      \rowcolor{lightblue!100} \quad $\Delta$ &
        \textcolor{red}{+4.1} &
        \textcolor{red}{+7.3} &
        \textcolor{red}{+6.0} &
        \textcolor{red}{+8.6} &
        \textcolor{red}{+6.5} \\
      \bottomrule
    \end{tabular}
  }
  \label{tab:main_baselines}
  \vspace{-8pt}
\end{table}

Tab.~\ref{tab:main_baselines} presents our main results on the LIBERO benchmark, comparing our TTT framework against state-of-the-art VLA models.
We apply TTT to the OpenVLA-OFT model (pre-trained with full trajectory demonstrations), enabling it to continue learning during deployment.

\textbf{Significant Performance Gains.}
Our TTT framework achieves substantial improvements across all four LIBERO task suites.
On average, we observe a \textbf{+6.5\%} absolute gain in success rate, elevating the baseline from 89.2\% to 95.8\%.
The improvements are consistent across diverse task types: +4.1\% on LIBERO-Spatial, +7.3\% on LIBERO-Object, +6.0\% on LIBERO-Goal, and most notably, \textbf{+8.6\%} on LIBERO-Long.
The substantial gain on LIBERO-Long is particularly significant, as this suite contains the most challenging long-horizon tasks with complex multi-step procedures.
With TTT, our method achieves 95.8\% average success rate, surpassing models like $\pi_0$ (94.2\%) and matching UniVLA (95.2\%), demonstrating that test-time adaptation can be as effective as collecting and training on large amounts of additional demonstration data.

\textbf{Challenge of Naive Reward Modeling.}
We also compare with SimpleVLA, which initially employs the binary outcome reward from the simulator, \ie, oracle reward.
We then replace the oracle reward with our progress estimator, and use a simple threshold-based approach to convert progress estimates into binary outcome rewards.
This version of SimpleVLA achieves only 87.7\% on LIBERO-Long, a modest +1.9\% improvement over the SFT-only baseline (85.8\%).
The limited gain highlights a critical challenge: \emph{directly using a noisy progress estimator to generate binary rewards for online RL is insufficient}.
In contrast, our accumulative progress estimation mechanism that smooths noisy signals and provides dense, stable feedback achieves 94.4\% (+8.6\%), demonstrating the importance of properly ``taming'' the reward model.

\subsection{TTT Under Low Data Regimes}

\begin{table}[!t]
  \centering
  \caption{Test-time training in low data regime, i.e., one demonstration for pre-training.}
  \resizebox{.6\linewidth}{!}{
    \begin{tabular}{lccccc}
      \toprule
      \multirow{2}{*}{\textbf{Model}} &
        \multicolumn{5}{c}{\textbf{LIBERO}} \\
      \cmidrule(lr){2-6}
      & \textbf{Spatial} & \textbf{Object} & \textbf{Goal} & \textbf{Long} & \textbf{Avg} \\
      \midrule
      OpenVLA-OFT         & 65.1 & 40.1 & 57.2 & 15.1 & 43.6 \\
      \textbf{EVOLVE-VLA}       & 73.4 & 70.0 & 64.7 & 37.1 & 61.3 \\
      \rowcolor{lightblue!100} \quad $\Delta$ &
        \textcolor{red}{+8.3} &
        \textcolor{red}{+29.9} &
        \textcolor{red}{+7.5} &
        \textcolor{red}{+22.0} &
        \textcolor{red}{+17.7} \\
      \bottomrule
    \end{tabular}
  }
  \label{tab:one_shot}
\end{table}

A key motivation for TTT is to reduce the labor cost of collecting extensive demonstration data. To evaluate TTT's effectiveness in low-data scenarios, we experiment with a more challenging setting: only \textit{one demonstration per task} for SFT pre-training\footnote{All 1-trajectory SFT models are reused from the SimpleVLA-RL released checkpoints~\cite{li2025simplevla}.}, followed by test-time training.

As shown in Tab.~\ref{tab:one_shot}, the 1-shot SFT baseline (OpenVLA-OFT) achieves only 43.6\% average success rate, indicating that a single demonstration is insufficient for learning robust manipulation policies.
However, applying our TTT framework yields substantial improvements, achieving 61.3\% average success rate, which is a remarkable \textbf{+17.7\%} absolute gain.
The improvements are consistent across all task suites: +8.3\% on LIBERO-Spatial, +29.9\% on LIBERO-Object, +7.5\% on LIBERO-Goal, and +22.0\% on LIBERO-Long.
These gains validate our core claim: test-time training can effectively alleviate the data collection burden by enabling learning from self-generated experiences rather than relying solely on extensive expert demonstrations.

\subsection{Toward Zero-Shot Cross-Task Generalization}

An intriguing capability enabled by our TTT framework is cross-task generalization through online learning.
To explore this, we conduct a preliminary experiment: we take a VLA model pre-trained exclusively on LIBERO-Long tasks (50 demonstrations per task) and directly deploy it on LIBERO-Object tasks without fine-tuning on task-specific demonstrations.

When deployed directly, the LIBERO-Long pre-trained policy achieves 0\% success rate on LIBERO-Object, as expected.
Although, conceptually, the two task suites may share some common motion primitives, the behavior cloning paradigm strongly hinders generalization.
Remarkably, however, by applying our TTT framework with progress-based feedback, the policy adapts purely through autonomous exploration and reaches \textbf{20.8\% success rate} on LIBERO-Object.
While this performance remains modest compared to task-specific SFT baselines (which achieve 40.1\% with a single demonstration and 96.6\% with 50 demonstrations), the ability to break 0 success rate without finetuning on task-specific human demonstrations represents a qualitatively different capability.
To the best of our knowledge, no prior VLA training method has demonstrated such cross-task transfer through test-time adaptation alone.
This preliminary result suggests that TTT, when paired with a foundation-level progress estimator like VLAC, can enable VLAs to generalize across task distributions through self-directed learning.

\begin{table}[!t]
  \centering
  \caption{Ablation study on accumulative progress estimation and and temporal sampling efficiency on LIBERO-Long task suite.}
  \resizebox{.7\linewidth}{!}{
    \begin{tabular}{lcccc}
      \toprule
      \textbf{Method} & \textbf{Sampling} & \textbf{Reward Calls} & \textbf{F-Score} & \textbf{SR (\%)} \\
      \midrule
      SFT & - & - & - & 85.8 \\
      Baseline (2 frames) & - & 32 & 0.04 & 88.3 \\
      \midrule
      Accumulative (4 frames) & Uniform & 96 & 0.09 & 90.1 \\
      Accumulative (8 frames) & Uniform & 224 & 0.17 & 89.3 \\
      \textbf{Accumulative (Ours)} & \textbf{Interval} & \textbf{32} & \textbf{0.20} & \textbf{91.3} \\
      \bottomrule
    \end{tabular}
  }
  \label{tab:abl_acc_sampling}
\end{table}

\subsection{Ablation Studies}

\paragraph{Accumulative Progress Estimation.}
Tab.~\ref{tab:abl_acc_sampling} validates the effectiveness and efficiency of our accumulative progress estimation mechanism with various frame sampling strategies.
F-Score is computed based on a balanced validation set (100 success cases, 100 failure cases) assessing task progress estimation performance.
The baseline that directly uses 2-frame critic values without accumulation achieves 88.3\% success rate with 32 reward calls but suffers from low F-score (0.04), indicating unreliable progress estimation.
When incorporating accumulative progress estimation, the sampling strategy for millstone frames matters.
The uniform sampling variants improve F-score over baseline but at the cost of significantly more reward calls (96 and 224 respectively), with diminishing or even negative returns in success rate,
suggesting that naive dense sampling introduces noise without proper temporal structure.
In contrast, our method achieves the best performance (91.3\% SR, 0.20 F-score) while maintaining computational efficiency with only 32 reward calls, demonstrating that interval-based sampling (\textbf{with a sliding $\Delta_{check}$}) combined with accumulative aggregation is both more effective and more efficient than naive uniform approaches.

\paragraph{Progressive Horizon Extension.}
Tab.~\ref{tab:abl_progressive} demonstrates the importance of progressive horizon extension for long-horizon tasks.
Starting from the SFT baseline (85.8\%), we examine three TTT variants.
First, using binary outcome rewards (thresholding the progress estimator) yields only 87.7\% (+1.9\%), confirming that converting dense progress into sparse signals loses valuable learning information.
Second, applying dense rewards from our accumulative progress estimator without progressive horizon achieves 91.3\% (+5.5\%), showing the benefit of dense feedback.
Finally, adding progressive horizon extension, \ie, gradually increasing the maximum rollout length during training, reaches 94.4\% (+8.6\%), providing an additional 3.1\% gain.
This validates our strategy: by initially constraining exploration to shorter horizons and progressively extending them, the policy learns more stable sub-task skills before tackling full-length trajectories, making it more resilient to estimation errors in long-horizon tasks.

\begin{table}[!t]
  \centering
  \caption{Ablation study on Progressive Horizon Extension with LIBERO-Long task suite.}
  \resizebox{0.65\linewidth}{!}{
    \begin{tabular}{lcc}
      \toprule
      \textbf{Method} & \textbf{Milestones} & \textbf{SR (\%)} \\
      \midrule
      Baseline (SFT only) & - & 85.8 \\
      \quad + Binary Outcome & - & 87.7 \\
      \quad + Dense Reward (Vanilla Critic) & - & 91.3 \\
      \quad + \textbf{Progressive Horizon} & \checkmark & \textbf{94.4} \\
      \bottomrule
    \end{tabular}
  }
  \label{tab:abl_progressive}
\end{table}

\subsection{Qualitative Analysis}
\label{sec:qualitative}

\begin{figure}[ht]
  \centering
  \includegraphics[width=0.75\linewidth]{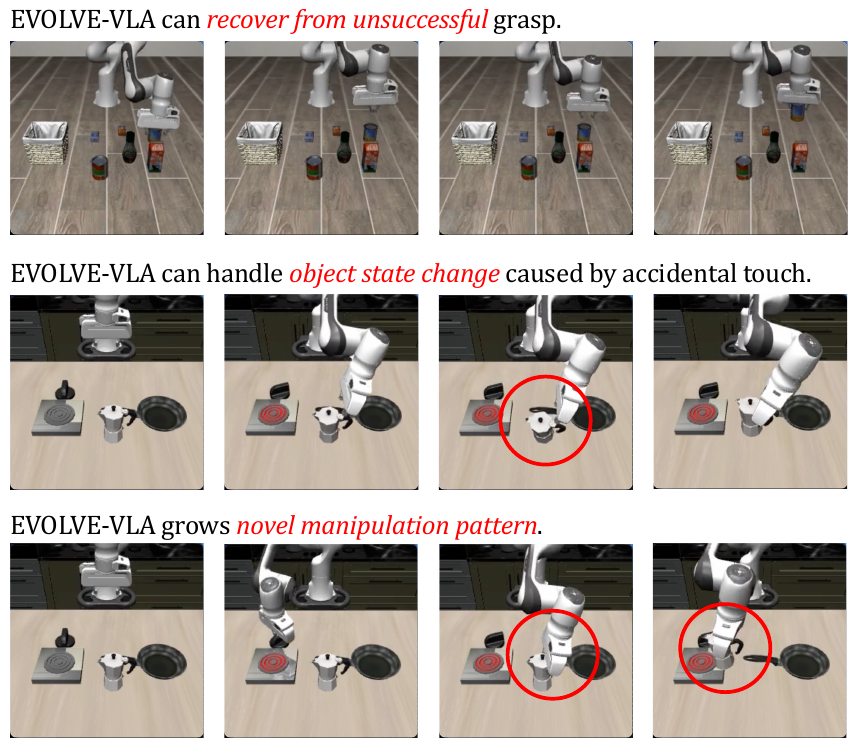}
  \caption{
  \textbf{Qualitative example of policy behavior with TTT.}
  } 
  \label{fig:qualitative}
\end{figure}

To gain deeper insights into how test-time training shapes policy behavior, we analyze representative rollout trajectories after TTT in Fig.~\ref{fig:qualitative}.
First, the policy develops error recovery capabilities: when initial grasp attempts fail, the SFT-only policy usually continue with the pre-programmed motion and fails, whereas after TTT the policy autonomously re-attempts grasping (top row).
Second, the policy adapts to pick an object but accidentally changed the object state, then it adjusts its motion to fit the new config rather than rigidly following memorized patterns (middle row).
Third, the policy discovers alternative manipulation strategies not present in demonstrations. For instance, grasping a pot by its body instead of the handle (bottom row).
These improvements indicate that progress-based feedback enables the policy to generalize beyond trajectory-level imitation to goal-oriented manipulation and explore diverse solutions.

\begin{figure}[ht]
  \centering
  \includegraphics[width=0.95\linewidth]{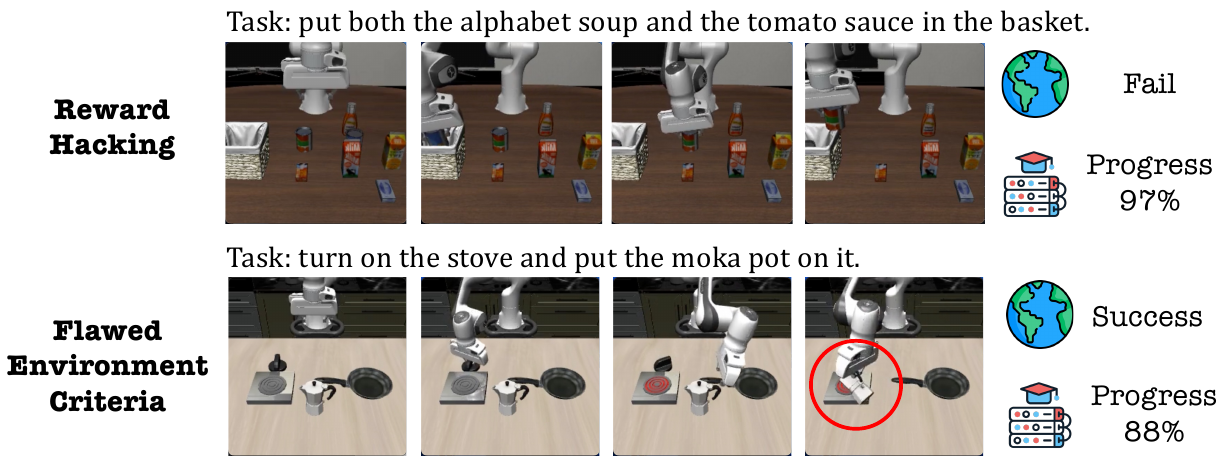}
  \caption{
  \textbf{Example of environment success criterion mismatch.}
  } 
  \label{fig:env_success_mismatch}
\end{figure}

Despite the improvements, we observe failure cases that reveal a fundamental challenge: misalignment between the environment's rule-based success criterion and the semantic task completion assessed by our progress estimator.
This mismatch manifests in two ways as shown in Fig.~\ref{fig:env_success_mismatch}.
First, in some cases the policy brings the scene very close to the goal state, leading the progress estimator to assign high rewards (near-completion signal), yet the environment's coordinate-based rules still judge the task as unsuccessful.
This creates a form of ``reward hacking'' where the policy optimizes for high progress scores without meeting the strict environmental criteria.
Second, the opposite occurs: the environment judges tasks as successful based on coordinate rules despite semantic incompleteness.
For instance, in Fig.~\ref{fig:env_success_mismatch}, a book placement task where the environment reports success because the book's coordinates satisfy the spatial constraints, yet semantically the book is not properly placed inside the shelf.
These misalignments highlight the inherent difficulty in aligning rule-based simulation criteria with semantic task understanding, suggesting that future work should explore improved calibration between progress estimators and environment oracles.

\section{Future Work}
Our work demonstrates that test-time training for VLA models is feasible by addressing noisy progress estimation through accumulative estimation and progressive horizon extension.
This foundational step enables VLAs to learn from experience rather than merely imitating demonstrations, unlocking several promising directions for future research.
First, developing more robust reward models would significantly enhance the framework's capabilities.
While our accumulative mechanism effectively handles noisy estimates, future reward models with better semantic alignment to environment success criteria could reduce the mismatch between progress estimation and rule-based success signals.
Additionally, improving zero-shot capability would eliminate the need for in-context examples, enabling truly zero-shot cross-task generalization.
While our current approach shows promising cross-task transfer (LIBERO-Long $\rightarrow$ LIBERO-Object), the reward model still benefits from task-specific context; future reward models trained on more diverse manipulation data, with better generalization capability could enable seamless adaptation to entirely novel tasks without any task-specific examples, even for the reward model.

Second, extending test-time training to real-world robotic deployment presents both opportunities and challenges.
The long \textbf{training times} required for online RL can be prohibitive in physical environments, where data collection is inherently slower than simulation.
Future work can explore techniques to accelerate real-world training, such as sim-to-real transfer for reward models, parallel robot deployment for distributed data collection, or more sample-efficient online learning algorithms.
Equally important is ensuring \textbf{safety} during exploration: the uncontrolled policy behavior in early training stages could damage the robot or environment.
Developing safety mechanisms—such as action constraints, safety critics, or human oversight protocols—would be crucial for enabling safe autonomous learning in physical environments.
Third, exploring more sophisticated exploration strategies and curriculum designs could further improve sample efficiency and enable adaptation to even more complex, long-horizon manipulation tasks.

\section{Conclusion}
We introduced EVOLVE-VLA, a test-time training framework that enables VLA models to continuously adapt through environment interaction, addressing the fundamental limitations of static SFT.
Inspired by how humans develop manipulation skills through practice and trial-and-error, our approach shifts VLAs from rigid trajectory memorization toward genuine adaptive learning.
By replacing impractical oracle rewards with a learned progress estimator and introducing two key technical contributions: (1) accumulative progress estimation and (2) progressive horizon extension, we demonstrate that VLAs can effectively learn from inherently noisy, self-generated feedback signals.
Our experiments on the LIBERO benchmark validate this approach, achieving +8.6\% on long-horizon tasks, +22.0\% in 1-shot learning, and enabling cross-task generalization (0\% → 20.8\%) without task-specific demonstration training.
Beyond these quantitative gains, we observe emergent capabilities like error recovery that arise purely from autonomous exploration.
We believe this work represents an essential step on the path toward truly general-purpose VLA systems that can continuously learn and improve in real-world deployment.

\printbibliography[heading=bibintoc]

\end{document}